\documentclass{article}
\usepackage{spconf,amsmath,graphicx,color,comment}
\usepackage{amsfonts}
\usepackage{amssymb,amsmath}
\usepackage{amsthm}
\usepackage{cite}
\usepackage{subcaption}
\usepackage{todonotes}
\usepackage{mathtools}
\usepackage{url}
\usepackage{multicol}
\usepackage{multirow}
\usepackage{subcaption}
\usepackage{midfloat}

\DeclareMathOperator{\Tr}{Tr} 
\title{Statistical Model-based Evaluation of Neural Networks} 
\name{Sandipan Das$^1$, Prakash B. Gohain$^1$, Alireza M. Javid$^1$, Yonina C. Eldar$^2$, Saikat Chatterjee$^1$\thanks{Thanks to Swedish Foundation for Strategic Research for their funding.}}
\address{$^1$KTH Royal Institute of Technology, Sweden\\
    $^2$Weizmann Institute of Science, Israel\\
    \{sandipan, pbg, almj, sach\}@kth.se, yonina.eldar@weizmann.ac.il} 

\newcommand{\sandipan}[1]{\textcolor{blue}{#1}}

\begin{document}

\maketitle
%================%
\begin{abstract}

Using a statistical model-based data generation, we develop an experimental setup for the evaluation of neural networks (NNs). The setup helps to benchmark a set of NNs vis-a-vis minimum-mean-square-error (MMSE) performance bounds. This allows us to test the effects of training data size, data dimension, data geometry, noise, and mismatch between training and testing conditions. In the proposed setup, we use a Gaussian mixture distribution to generate data for training and testing a set of competing NNs. Our experiments show the importance of understanding the type and statistical conditions of data for appropriate application and design of NNs.

%(GMM) for  based data and compute the MMSE bounds exactly. Experiments show that a trade-off between complexity and performance is important in the choice of a neural network. We find that a readily available complex neural network may suffer in performance saturation under favorable conditions. The experiments underscore the importance of domain knowledge in the design and choice of a neural network for a given dataset. 

%moderately complex neural networks are good. They are robust when we consider noisy data and limited training data. We also show that the celebrated \sandipan{??} ResNet - the most complex method that we evaluate in experiments - suffers in performance saturation under favourable conditions. \sandipan{It's a very strong claim IMHO!}

\end{abstract}

%=====================%
\section{Introduction}
%=====================%
% Neural networks (NNs) have received much attention in the last decade for their success in common inference tasks, such as estimation, prediction, regression and pattern classification. NNs are typically model-free, which means they do not try to incorporate explicit models or physical laws explaining data generation procedures. A well known and widely used NN is the feed-forward NN which is applied for many kinds of data in varying applications. In a feed-forward NN, an input signal is passed through a series of linear and non-linear transforms to provide an output. Non-linear transforms are known as activation functions in NN literature. The parameters, for example, the linear transforms (weight matrices) of a NN is learned using a training dataset. This is perhaps the main reason why NNs are often called data-driven systems. 

%Neural networks (NNs) and deep learning are crucial for modern machine learning (ML) and data science. 
It is imperative to understand and explain neural networks (NNs) for %{improvements of ML and} 
realizing their wide scale use in critical applications. Explaining decisions of (deep) NNs remains as an open research problem \cite{Sam19}, \cite{rieger2020aggregating}.
%We use ‘interpretability’ and ‘explainability’ interchangeably in this article. 
%Through literature survey on explainable machine learning, our understanding is that e
Existing methods for explanations can be categorized in various ways, \cite{lipton2017mythos}, 
%However, there is no consensus yet \cite{lipton2017mythos}.
%on categorization in the current literature.
for example, \emph{intrinsic} and \emph{post-hoc} \cite{post-hoc-ribeiro}. Intrinsic interpretability is inherent in structures of methods, such as rule-based models, decision trees, linear models, and unfolding NNs \cite{monga2019algorithm}. In contrast, a post-hoc method requires a second (simple) model to provide explanations about the original (complex) method. There are approaches to provide
\emph{local} and \emph{global} explanations. A local explanation justifies a NN's output for a specific input. A global explanation provides a justification on average performance of a NN, independently of any particular input \cite{danilevsky2020survey}.

Techniques for explainability include visualization of neuronal activity in layers of a deep architecture, for examples, based on sparsity and heatmaps \cite{Sam19, yosinski-2015-ICML-DL-understanding-neural-networks, raghakotkerasvis, cadena2018diverse, zhou2016cvpr, wagner2019interpretable}. There are also techniques for visualizing training steps of non-convex optimization methods (such as stochastic gradient search), and identifying contributions of input features, referred to as feature importance \cite{lemhadri2020lassonet}. Example-driven techniques explain an output for a given input by identifying and presenting other instances, usually from available labeled data, that are semantically similar to the input instance. This also includes adversarial perturbation-based studies for explaining robustness \cite{moosavidezfooli2016deepfool}. There is a recent
trend to develop model-based design and analysis for explanations. In particular, the work \cite{rieger2020aggregating} trains deep NNs using statistical and domain specific model-based perturbations.  

\noindent\emph{Our Contributions:} We develop an experimental setup to evaluate and compare NNs. The proposed setup is based on traditional statistical signal processing. We use generated data from tractable probability distributions that has computable minimum-mean-square-error (MMSE) estimator. The MMSE estimation performance can be used as a bound to compare the chosen NNs. While sophisticated generative models exist in literature, such as normalizing flows \cite{normalizing_flows_2020}, we choose a generative model (probability distribution) for which
the MMSE estimator can be computed in a simple form. Therefore we use Gaussian mixture model (GMM) for the proposed setup. In theory, a GMM can model arbitrary distributions closely by increasing number of mixture components \cite[Chapter 3]{Goodfellow-et-al-2016}. The MMSE estimator for joint GMM distribution has a closed form and interpretable expression \cite{Kundu2008GMMDomain}. Availability of the closed form MMSE expression is beneficial to perform controlled experiments. In addition, GMM helps in easy visualization of data geometry. Our setup helps in studying the performance trend of a set of NNs under different conditions like: training data need, data dimension, data geometry and effects of different testing conditions, and provides a view on the probable failure cases of NNs.

The rest of the paper is organised as follows. Section 2 presents the statistical model-based experimental setup. The data generation process, experiments and results are presented in Section 3. Finally Section 4 provides conclusions.

%=========================================
\section{Statistical model-based \\ experimental setup} 
%=========================================
A typical setup in statistical signal processing is estimation of a target signal $\mathbf{t} \in \mathbb{R}^Q$ given an observation signal $\mathbf{x} \in \mathbb{R}^P$. Denoting the estimated signal as $\tilde{\mathbf{t}}$,
%Using MSE $\mathcal{E}\{ \| \mathbf{t} - \tilde{\mathbf{t}} \|^2 \}$, the minimum MSE (MMSE) \cite{Kay1993FundamentalsTheory} estimator is 
the MMSE estimator is defined as \cite{Kay1993FundamentalsTheory}
\begin{eqnarray}
\tilde{\mathbf{t}}_{\text{MMSE}} = \arg \min_{\tilde{\mathbf{t}}}  \mathcal{E}\{ \| \mathbf{t} - \tilde{\mathbf{t}} \|^2 \} = \mathcal{E}(\mathbf{t}|\mathbf{x}),
\end{eqnarray}
and the MMSE performance is computed as
\begin{eqnarray}
C_{\text{MMSE}} = \mathcal{E}\{ \| \mathbf{t} - \mathcal{E}(\mathbf{t}|\mathbf{x}) \|^2 \}.
\end{eqnarray}
The above MMSE performance provides a theoretical bound. Therefore we can compare performance of a set of NNs with the MMSE bound. If the joint distribution $p(\mathbf{x},\mathbf{t})$ is parametarized by $\boldsymbol{\theta}$, then we can perform different controlled experiments by varying $\boldsymbol{\theta}$ followed by computing $C_{\text{MMSE}}$ and checking how the chosen NN fares against the MMSE. These experiments will provide an analysis platform for understanding the conditions of success and failure of a NN. 
However, in delineating such a path of analysis we encounter a technical challenge: there exist few distributions for which MMSE estimators are computable analytically and we can evaluate $C_{\text{MMSE}}$ reliably. The next subsection presents one such distribution and a helpful signal model for conducting our controlled experiments. 

\subsection{Distributions, system model and their advantages}
%=======================================%
%\section{System model and optimal MMSE estimator}
%=======================================%
If the joint distribution $p(\mathbf{x},\mathbf{t})$ has a Gaussian mixture (GM) density with parameter $\boldsymbol{\theta}$ then we can compute the optimal (non-linear) MMSE estimator $\tilde{\mathbf{t}}_{\text{MMSE}} = \mathcal{E}_{\boldsymbol{\theta}}(\mathbf{t}|\mathbf{x})$ exactly in an analytical form. Using simulations and invoking the law of large numbers, we then can evaluate the MMSE performance.
We use a classical linear system model (observation system) for the controlled experiments as shown below
\begin{eqnarray}
\mathbf{x} = \mathbf{Ht} + \mathbf{n}, \label{eq:linear_model}
\end{eqnarray}
where $\mathbf{x} \in \mathbb{R}^P$ is the observation signal,  $\mathbf{t} \in \mathbb{R}^Q$ is the target signal vector to be estimated and $\mathbf{n} \in \mathbb{R}^P$ is the observation noise. The matrix $\mathbf{H} \in \mathbb{R}^{P\times Q}$ represents a system which is assumed to be known. The fundamental advantages of using a linear model are as follows:
(a) For the linear observation system, if $\mathbf{t}$ is GM distributed and $\mathbf{n}$ is Gaussian distributed then the joint distribution $p(\mathbf{x},\mathbf{t})$ is also a GM density. This enables us to compute the MMSE performance. 
(b) The use of GM distribution can assist in generating fairly complicated (data spread) data geometry, which in turn can be used to benchmark a NN against the MMSE for different data geometries. 
%(3) The linear observation model is a generative model and we can create required amount of data to train a NN. This provides understanding about the requirement of training data amount. 
(c) The linear observation model is a generative model and as such we can create any required amount of training and testing data. This enables us to visualize the effect of training data on NN performance.
%This provides understanding about the requirement of training data amount. 
(d) We can compute the signal-to-noise-ratio (SNR) at varying conditions and check the debilitating effect of noise on the NN. 
(e) We can check the role of observation's dimension 
by varying $P$.
We consider the signal $\mathbf{t}$ to be GM distributed, and noise $\mathbf{n}$ to be Gaussian distributed as follows
\begin{equation}
p(\mathbf{t}) = \sum_{m=1}^{M}\alpha_{m}\mathcal{N}(\mathbf{t};\boldsymbol{\mu}_m,\mathbf{C}_m) ;\quad 
p(\mathbf{n}) = \mathcal{N}(\boldsymbol{\mu}_{\mathbf{n}},\mathbf{C}_{\mathbf{n}}), 
\label{eq:GM_source_and_Gaussian_noise}
\end{equation}

\begin{figure*}
\hrule
\begin{eqnarray}
\label{eq:MMSE_estimator}
\begin{array}{c}
 \tilde{\mathbf{t}}_{\text{MMSE}} = \mathcal{E}(\mathbf{t}|\mathbf{x}) = \displaystyle\sum_{m=1}^{M} \beta_{m}(\mathbf{x})\left(\pmb{\mu}_{m}+\mathbf{C}_{m} \mathbf{H}^{T}\left[\mathbf{H} \mathbf{C}_{m} \mathbf{H}^{T}+\mathbf{C}_{\mathbf{n}}\right]^{-1}\left(\mathbf{x}-\left[\mathbf{H} \pmb{\mu}_{m}+\pmb{\mu}_{\mathbf{n}}\right]\right)\right) ,  
\\
\beta_{m}(\mathbf{x}) =\displaystyle\frac{\alpha_{m} \frac{1}{\left(\sqrt{2 \pi}\right)^{P}\left|\mathbf{H} \mathbf{C}_{m} \mathbf{H}^{T}+\mathbf{C}_{\mathbf{n}}\right|^{\frac{1}{2}}} \exp \left[-\frac{1}{2}\left(\mathbf{x}-\left[\mathbf{H} \pmb{\mu}_{m}+\mu_{\mathbf{n}}\right]\right)^{T}\left[\mathbf{H} \mathbf{C}_{m} \mathbf{H}^{T}+\mathbf{C}_{\mathbf{n}}\right]^{-1}\left(\mathbf{x}-\left[\mathbf{H}\pmb{\mu}_{m}+\mu_{\mathbf{n}}\right]\right)\right]}{\sum_{j=1}^{M} \alpha_{j} \frac{1}{\left(\sqrt{2 \pi}\right)^{P}\left|{\mathbf{H}} \mathbf{C}_{j} \mathbf{H}^{T}+\mathbf{C}_{\mathbf{n}}\right|^{\frac{1}{2}}} \exp \left[-\frac{1}{2}\left(\mathbf{x}-\left[\mathbf{H} \pmb{\mu}_{j}+\mu_{\mathbf{n}}\right]\right)^{T}\left[\mathbf{H} \mathbf{C}_{j} \mathbf{H}^{T}+\mathbf{C}_{\mathbf{n}}\right]^{-1}\left(\mathbf{x}-\left[\mathbf{H}\pmb{\mu}_{j}+\mu_{\mathbf{n}}\right]\right)\right]}. 
\end{array}
\end{eqnarray}
\hrule
\end{figure*}

% \begin{eqnarray}
% \begin{array}{c}
% p(\mathbf{t}) = \displaystyle\sum_{m=1}^M \alpha_m \, \mathcal{N}(\mathbf{t}; \pmb{\mu}_m, \mathbf{C}_m), \\  \\
% p(\mathbf{n}) = \mathcal{N}(\pmb{\mu}_{\mathbf{n}},\mathbf{C}_{\mathbf{n}}).
% \end{array}
% \label{eq:GM_source_and_Gaussian_noise}
% \end{eqnarray}
\noindent where
$M$ is the number of Gaussian mixture components, $\alpha_m$ is the mixing proportion subject to $\sum_{m=1}^M \alpha_m = 1$,  $\boldsymbol{\mu}_m$ and $\mathbf{C}_m$ are the mean and covariance of the $m^{th}$ Gaussian distribution respectively. Furthermore, $\boldsymbol{\mu}_{\mathbf{n}}$ and $\mathbf{C}_{\mathbf{n}}$ are the mean and covariance of noise respectively. The joint probability density $p(\mathbf{x},\mathbf{t})$ is also GM distributed with parameter $\boldsymbol{\theta} = \{ \{\alpha_m,\boldsymbol{\mu}_m, \mathbf{C}_m\}_{m=1}^M, \boldsymbol{\mu}_{\mathbf{n}}, \mathbf{C}_{\mathbf{n}} \}$, and the explicit analytical form of the MMSE estimator is shown in \eqref{eq:MMSE_estimator} (as per \cite{Kundu2008GMMDomain}). 
%Note that the MMSE estimator is an weighted linear combination of linear estimators where the weights $\{ \beta_m(\mathbf{x}) \}_{m=1}^M$ are non-linear function of $\mathbf{x}$.

While we have an analytical expression for $\tilde{\mathbf{t}}_{\text{MMSE}}$, we do not have an analytical expression for the MMSE $\mathcal{E}\{\| \mathbf{t}- \tilde{\mathbf{t}}_{\text{MMSE}} \|^2\}$. The MMSE is thus computed through simulations by averaging the $l_2$ norm of the estimation error over a large number of samples. 
\begin{comment}
% In addition to the MMSE estimator, we can also benchmark the NNs with the optimal linear estimator (strictly speaking the affine transform). The linear MMSE (LMMSE) estimator and its performance have analytical expressions, shown in \eqref{eq:LMMSE_estimator}.
\begin{strip}
\begin{eqnarray}
\label{eq:LMMSE_estimator}
\begin{array}{c}
\tilde{\mathbf{t}}_{\text{LMMSE}} =  \boldsymbol{\mu}_{\mathbf{t}}+\mathbf{C}_{\mathbf{t}} \mathbf{H}^{T}\left(\mathbf{H} \mathbf{C}_{\mathbf{t}} \mathbf{H}^{T}+\mathbf{C}_{\mathbf{n}}\right)^{-1}\left(\mathbf{x}-\mathbf{H} \boldsymbol{\mu}_{\mathbf{t}}\right),\\ \\
\mathcal{E}\{\| \mathbf{t}- \tilde{\mathbf{t}}_{\text{LMMSE}} \|^2\} =  trace\left[\mathbf{C}_{\mathbf{t}}-\mathbf{C}_{\mathbf{t}} \mathbf{H}^{T}\left(\mathbf{H} \mathbf{C}_\mathbf{t} \mathbf{H}^{T}+\mathbf{C}_\mathbf{n}\right)^{-1} \mathbf{H} \mathbf{C}_\mathbf{t}\right].
\end{array}
\end{eqnarray}
\end{strip}
where $\pmb{\mu}_{\mathbf{t}}$ and $\mathbf{C}_{\mathbf{t}}$ denote mean and covariance of target signal $\mathbf{t}$ as
\begin{eqnarray}
\begin{array}{c}
\pmb{\mu}_{\mathbf{t}} \triangleq \mathcal{E}\{\mathbf{t}\} = \displaystyle\sum_{m=1}^M \alpha_m \pmb{\mu}_m, \\ 
\begin{array}{rl}
\mathbf{C}_{\mathbf{t}}     & =  \mathcal{E}\{ (\mathbf{t} - \mathcal{E}(\mathbf{t})) (\mathbf{t} - \mathcal{E}(\mathbf{t}))^T  \} \\
     & 
\end{array}
\end{array}
\end{eqnarray}
\end{comment}
We use the normalized mean-square-error (NMSE) in decibel (dB) scale as the measure of performance for our experiments, as follows:
%which is shown below to compare the experiments in similar scale, %\sandipan{Saikat: Please check if this similar scale statement is correct or not? Saikat: People in this community are aware of the samle scale thing.}
$
%\begin{equation}
\mathrm{NMMSE}_{\text{dB}} = 10 \log_{10} \frac{\mathcal{E}\{\| \mathbf{t} - \tilde{\mathbf{t}} \|^2 \}}{\mathcal{E}\{\| \mathbf{t} - \mathcal{E}(\mathbf{t}) \|^2 \} }.
%\label{eq:NMMSE_def}
%\end{equation}
$
Here, $\mathcal{E}\{\| \mathbf{t} - \tilde{\mathbf{t}} \|^2 \}$ is the empirical estimation error power and $\mathcal{E}\{\| \mathbf{t} - \mathcal{E}(\mathbf{t}) \|^2 \}$ is the signal power computed analytically.
% \begin{eqnarray}
% \resizebox{0.93 \columnwidth}{!}{$
% \begin{array}{rl}
% \mathcal{E}\{\| \mathbf{t} - \mathcal{E}(\mathbf{t}) \|^2 \} & = \mathrm{trace}[\mathbf{C}_{\mathbf{t}}] \\
% & = \mathcal{E}\{ \mathrm{trace}[(\mathbf{t} - \mathcal{E}(\mathbf{t})) (\mathbf{t} - \mathcal{E}(\mathbf{t}))^T]  \} \\
% & = \mathrm{trace}[\mathcal{E}\{ (\mathbf{t} - \mathcal{E}(\mathbf{t})) (\mathbf{t} - \mathcal{E}(\mathbf{t}))^T  \}] \\
% & = \mathrm{trace}[\mathcal{E}\{ \mathbf{t} \mathbf{t}^T \}] - \mathrm{trace}[ \mathcal{E} (\mathbf{t})[\mathcal{E}(\mathbf{t})]^T ]  \\
% & = \mathrm{trace}[ \int_{\mathbf{t}}\mathbf{t}\mathbf{t}^T p(\mathbf{t}) \, d\mathbf{t} ] - \mathrm{trace}[ \mathcal{E}(\mathbf{t})[\mathcal{E}(\mathbf{t})]^T ]
% \end{array}
%  $}
% \end{eqnarray}
\begin{comment}
\begin{eqnarray}
\resizebox{0.93 \columnwidth}{!}{$
\begin{array}{rl}
\mathcal{E}\{\| \mathbf{t} - \mathcal{E}(\mathbf{t}) \|^2 \} & = \Tr[\mathbf{C}_{\mathbf{t}}] = \mathcal{E}\{\Tr[(\mathbf{t} - \mathcal{E}(\mathbf{t})) (\mathbf{t} - \mathcal{E}(\mathbf{t}))^T]  \} \\
& = \Tr\left[ \int_{\mathbf{t}}\mathbf{t}\mathbf{t}^T p(\mathbf{t}) \, d\mathbf{t} ] - \Tr[ \mathcal{E}(\mathbf{t})[\mathcal{E}(\mathbf{t})]^T \right]
\end{array}
 $}\label{eq:signal_power}
\end{eqnarray}
\end{comment}

%=====================================%
\section{Generation of Data and Experiments}
%==========================================%
In this section, we present the data generation procedure and the experiments performed for different data geometries. 
%We use training and test data-sets to train and test the NNs. 

%{\textcolor{red}{We will compare between theoretical estimators and training based estimators. Theoretical estimators: Optimal MMSE and Optimal linear MMSE (LMMSE). Training based estimators: Linear estimator and non-linear estimators. Non-linear estimators can be Kernel estimator, SSFN, ELM. Now comes one simulation with fixed conditions, where you can compare with famous neural nets - VGG, U-Net, ResNet, CNN, DenseNet, etc etc.
%}}
%----------------------------%
\subsection{Data generation}
\begin{figure}
      \centering
      \includegraphics[width=1.0\linewidth,scale=1.0]{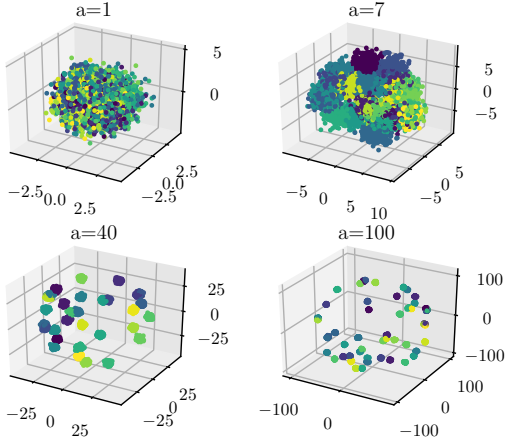}
      \captionof{figure}{Visualization of GM data geometries using  \eqref{eq:GM_source_and_Gaussian_noise_scaled}. As $a$ increases the inter-component separability increases.}
      \label{fig:exp_torroid_data_3d}
\end{figure}

%----------------------------%
A GM is a universal approximator of densities which can approximate any smooth density with enough numbers of mixture components \cite[Chapter 3]{Goodfellow-et-al-2016}. We generate our training data from \eqref{eq:linear_model} by sampling from the following GMM,
%\textcolor{red}{The following settings are assumed for the GM distribution given in \eqref{eq:GM_source_and_Gaussian_noise}
%:$\alpha_m = 1/M$, $\|\boldsymbol{\mu}_m \|^2 = 1$, $\mathbf{C}_m = \mathbf{I}_Q$, $\boldsymbol{\mu}_{\mathbf{n}} = \mathbf{0}$, $\mathbf{C}_{\mathbf{n}} = \frac{b}{P} \mathbf{I}_P$ (maybe we can make this concise)}
%\sandipan{It is not saving too much space IMHO. Commenting. But let's discuss more}
\begin{eqnarray}
\begin{array}{c}
p(\mathbf{t}) = \displaystyle\sum_{m=1}^M \alpha_m \, \mathcal{N}(\mathbf{t}; a\pmb{\mu}_m, \mathbf{C}_m), \\ 
p(\mathbf{n}) = \mathcal{N}\left(\pmb{\mu}_{\mathbf{n}},\mathbf{C}_{\mathbf{n}}\right) = \mathcal{N}\left(\mathbf{0},\frac{b}{P}\mathbf{I}_P\right),
\end{array}
\label{eq:GM_source_and_Gaussian_noise_scaled}
\end{eqnarray}
where $a$ and $b$ are the parameters which helps in generating different data geometries and signal-to-noise-ratio (SNR) $\frac{\mathcal{E}\{\| \mathbf{t} - \mathcal{E}(\mathbf{t}) \|^2 \}}{\mathcal{E}\{\| \mathbf{n} \|^2 \}}$. For experimental analysis, we can create varying data geometries by choice of $a$. Assuming $\|\boldsymbol{\mu}_m \|^2=1$, the choice of $a$ provides position of the mean vectors of the Gaussian components on a $Q$-dimensional sphere with radius $a$. Therefore, by varying $a$ we can represent a gamut of distributions from a single mode to multiple modes. If we choose $a=0$, then all the components superimpose on each other and the GM becomes a single mode Gaussian distribution. If we increase $a$ separation between the components increases, and also increases the signal power \eqref{eq:Signal_power_1}.
A pictorial view of the generated data geometries (projected in 3-D) by increasing $a$, but keeping the same covariances across mixtures, are shown in Fig.~\ref{fig:exp_torroid_data_3d}.
%\textcolor{red}{(what does appropriate mean? It is not clear! Data geometry ((modes of distribution))!).} 
%In this case the MMSE estimator is shown in \eqref{eq:MMSE_Estimator_TorroidData}.
For ease of experimentation, let 
$\alpha_m = \frac{1}{M}, \| \pmb{\mu}_m \|^2 =1, \mathbf{C}_m = \mathbf{I}_Q.$
The MMSE estimator for the observation model  \eqref{eq:linear_model} can be found using \eqref{eq:MMSE_estimator} by substituting the chosen set of parameters.
% \begin{figure}
%       \centering
%       \includegraphics[width=\linewidth,scale=1.0]{plots/A/torrid2d.eps}
%       \caption{2-D torrid data for different values of $a$}
%       \label{fig:exp_torroid_data_2d}
% \end{figure}

% This is the figure Saikat commented to avoid long compile time. Later we have to uncomment the figure
%\begin{comment}
%\end{comment}

%Note that the covariance of all the GMs are fixed to $\mathbf{I}_Q$ and the mean of the $m$'th GM is $a\pmb{\mu}_m$ with the constraint $\| \pmb{\mu}_m \|^2 =1$.
%\textcolor{blue}{If we choose $a=0$ then all the GMs superimpose on each other and the GM distribution boils down to a Gaussian distribution. On the other hand, if we increase $a$ then the mean vectors of the GMs are on a sphere whose radius is increasing. This leads to more separation between the mixture components. }\\
\noindent The signal power $\mathcal{E}\{\| (\mathbf{t} - \mathcal{E}(\mathbf{t}) \|^2 \}$ is shown in \eqref{eq:Signal_power_1} as, 
\begin{eqnarray}
\resizebox{0.92 \columnwidth}{!}{$
\begin{array}{rl}
\label{eq:Signal_power_1}
&\mathcal{E}\{\| (\mathbf{t} - \mathcal{E}(\mathbf{t}) \|^2 \} \\ &= \Tr\left[ \displaystyle\int_{\mathbf{t}}\mathbf{t}\mathbf{t}^T p(\mathbf{t}) \, d\mathbf{t} \right] - \Tr[ \mathcal{E}(\mathbf{t})[\mathcal{E}(\mathbf{t})]^T ] \\
 &= Q + a^2 \left( 1 -  \frac{1}{M^2} \Tr \left[[\displaystyle\sum_{m=1}^M \pmb{\mu}_m] [\sum_{m=1}^M \pmb{\mu}_m]^T \right] \right).
\end{array}
 $}
 \end{eqnarray}
The noise power is $\mathcal{E}\{\| (\mathbf{n} - \mathcal{E}(\mathbf{n}) \|^2 \} = \sum_{i=1}^P \mathcal{E} (n_i^2) = P \frac{b}{P} = b$.
%\noindent 
The signal-to-noise-ratio (SNR) in dB is,
\begin{alignat}{1}
\label{eq:SNR_GM_source_and_Gaussian_noise_scaled}
& SNR = 10 \log_{10} \frac{\mathcal{E}\{\| (\mathbf{t} - \mathcal{E}(\mathbf{t}) \|^2 \}}{\mathcal{E}\{\| (\mathbf{n} - \mathcal{E}(\mathbf{n}) \|^2 \}} \\
& = 10 \log_{10} \frac{Q + a^2 \left( 1 -  \frac{1}{M^2} \Tr [\sum_{m=1}^M \pmb{\mu}_m] [\sum_{m=1}^M \pmb{\mu}_m]^T \right)}{b}. \nonumber
\end{alignat}

For the data generation, we sample $\mathbf{t}$ and $\mathbf{n}$ from the distributions \eqref{eq:GM_source_and_Gaussian_noise}, and generate $\mathbf{x}$ using \eqref{eq:linear_model}. We denote the training set as $\mathcal{D}_{train} = \{ (\mathbf{x}^{(j)}, \mathbf{t}^{(j)}) \}_{j=1}^{J_{train}}$ with $J_{train}$ data-and-target pairs and the test set as $\mathcal{D}_{test} = \{ (\mathbf{x}^{(j)}, \mathbf{t}^{(j)}) \}_{j=1}^{J_{test}}$ with $J_{test}$ data-and-target pairs. The experiments are carried out in Monte-Carlo fashion over many realizations of observation system matrix $\mathbf{H}$ and the results are shown averaged over those realizations. The components of $\mathbf{H}$ are i.i.d Gaussian.

\begin{comment}
% Save space

\begin{figure*}
\hrule
\begin{eqnarray}
\label{eq:MMSE_Estimator_TorroidData}
\begin{array}{c}
 \tilde{\mathbf{t}}_{MMSE}  =\displaystyle\sum_{m=1}^{M} \beta_{m}(\mathbf{x})\left(a\pmb{\mu}_{m}+\mathbf{C}_{m} \mathbf{H}^{T}\left[\mathbf{H} \mathbf{C}_{m} \mathbf{H}^{T}+\frac{b}{P}\mathbf{I}_P \right]^{-1} \left(\mathbf{x}-a \mathbf{H}\pmb{\mu}_{m}\right)\right),  \\
\beta_{m}(\mathbf{x}) = \displaystyle\frac{\alpha_{m} \frac{1}{\left(\sqrt{2 \pi}\right)^{P}\left|\mathbf{H} \mathbf{C}_{m} \mathbf{H}^{T}+\frac{b}{P}\mathbf{I}_P\right|^{\frac{1}{2}}} \exp \left[-\frac{1}{2}\left(\mathbf{x}-a\mathbf{H}  \pmb{\mu}_{m}\right)^{T}\left[\mathbf{H} \mathbf{C}_{m} \mathbf{H}^{T}+\frac{b}{P}\mathbf{I}_P\right]^{-1}\left(\mathbf{x}-a\mathbf{H} \pmb{\mu}_{m}\right)\right]}{\sum_{j=1}^{M} \alpha_{j} \frac{1}{\left(\sqrt{2 \pi}\right)^{P}\left|{\mathbf{H}} \mathbf{C}_{j} \mathbf{H}^{T}+\frac{b}{P}\mathbf{I}_P\right|^{\frac{1}{2}}} \exp \left[-\frac{1}{2}\left(\mathbf{x}-a\mathbf{H} \pmb{\mu}_{j}\right)^{T}\left[\mathbf{H} \mathbf{C}_{j} \mathbf{H}^{T}+\frac{b}{P}\mathbf{I}_P\right]^{-1}\left(\mathbf{x}-a\mathbf{H} \pmb{\mu}_{j}\right)\right]}
\end{array}
\end{eqnarray}
\hrule
\end{figure*}

\end{comment}

\subsection{Experiments}
\label{subsec:EstimationExperiments_DataOverRing}

%\begin{enumerate}
%    \item {\textcolor{blue}{Task 1: Can you plot Figures \ref{fig:exp_2_a10}, \ref{fig:exp_1} and \ref{fig:exp_5} in the same scale. So that we can compare between the figures. I think the x-axis SNR can be in the range -5 - 35 dB.}}
%    \item {\textcolor{blue}{Task 2: Can you implement `Kernel substitution' and CNN plots in the figures. If possible ResNet too.}}
    %\item {\textcolor{blue}{Task 3: Can you use ELM and SSFN with 500 nodes / layer. So that nobody can say that we used small width.}}
    %\item {\textcolor{blue}{Task 4: Can you repeat the experiments using 2000 data points instead of 1000 data points that you have currently used.}}
%\end{enumerate}
%We now perform experiments to understand several aspects.
Experiments are performed to understand several aspects of the NNs. We choose four types of NNs: extreme learning machine (ELM) \cite{Huang06extremelearning}, self-size estimating feedforward NN (SSFN) \cite{2019arXiv_ssfn}, feedforward NN (FFNN) \cite{HORNIK1989_ffnn}, and residual network (ResNet) \cite{he2015resnet}. The choice of NNs is based on their structures and training complexity. 

We choose an ELM with a single hidden layer comprised of 30 hidden nodes. The ELM uses a random weight matrix. Therefore ELM is a simple network to train. The SSFN is a multi-layer network with low training complexity. The SSFN uses a combination of random weights and learned parameters. We used a 20-layer SSFN with 30 hidden nodes per layer. Then we evaluate a FFNN with 6 hidden layers. The number of nodes are 64, 128, 256, 256, 128, 64 for the six layers, respectively. FFNN is trained using standard backpropagation. Finally we use ResNet-34 to have a deep NN that has capability to avoid the vanishing gradient problem \cite{he2015resnet}. The ResNet-34 is a 34 layered architecture with a combination of convolutional layers, max-pooling, batch normalization and skip connections. We adapted it from Keras ResNet-50 structure with one change - use of 1-D convolution instead of 2-D convolution because of our data type. The number of trainable parameters are approximately 0.3K, 6K, 240K, and 500K for ELM, SSFN, FFNN, and ResNet-34, respectively.
All the experiments were performed with a train:test split of $70:30$ using $10$ Monte Carlo simulations.
%The experiments will be conducted in `\emph{matched training and testing statistics scenario}'.

\begin{figure}[t]
	\centering
	\includegraphics[trim={0cm 0cm 0cm 0cm},clip,scale = 0.5]{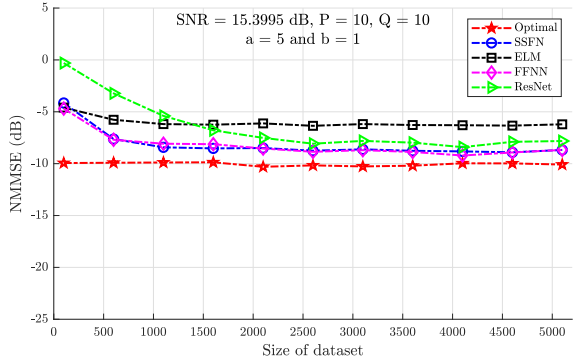}
	\caption{Estimation performance study for visualizing the effect of training data amount.}
	\label{fig:exp_4}
\end{figure}

Our first three experiments are conducted in \emph{matched training and testing statistics scenario}, which means we first fix the parameters of distributions to define the signal statistics and then generate $\mathcal{D}_{train}$ and $\mathcal{D}_{test}$.
In our first experiment, we study the effect of training dataset size $\mathcal{D}_{train}$. The estimation performance versus size of training dataset is shown in Fig.~\ref{fig:exp_4}. The experiment shows that performances of all the four NNs saturate as the training data size increases. Therefore the popular idea that more and more training data improves the learning capacity of NNs is questionable. Based on the saturation trend, we decide to perform rest of the experiments using 3000 total data.
% \begin{Figure}[t]
%       \includegraphics[width=1.0\textwidth]{plots/A/mmse_4.eps}
%         \captionof{figure}{Estimation performance study for visualizing the effect of training data amount.} %Here we evaluated the NMMSE for the different networks for increasing cardinality $\mathcal{D}_{train}$ of training dataset. Here we have fixed $b$ and $a$, i.e. the signal statistics are fixed for all the different scenarios.
%         \label{fig:exp_4}
% \end{Figure}

\begin{figure}[t]
	\centering
	\includegraphics[trim={0cm 0cm 0cm 0cm},clip,scale = 0.5]{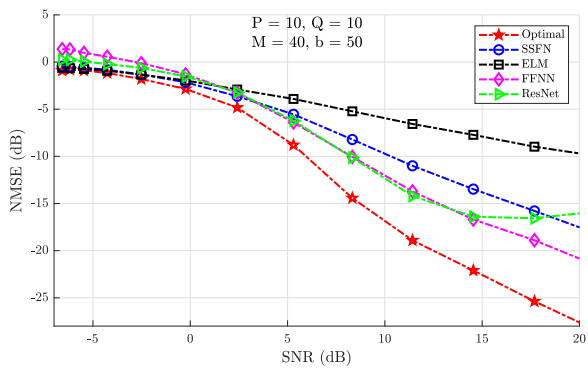}
	\caption{Estimation performance study for visualizing role of data geometry and the resulting signal power.}
	\label{fig:exp_1}
\end{figure}

The second experiment is designed to study how the data geometry %\textcolor{red}{(What do we mean by signal geometry?Not very clear.)} 
affects the estimation performance. The choice $b=50$ translates to a fixed noise power. The SNR is varied by varying $a$. %The estimation performances versus SNR is shown in Figure~\ref{fig:exp_1}. 
%\textcolor{red}{Note that on increasing the value of $a$ the signal power increases in addition the distance between Gaussian component mean vectors also increases. When $a=0$, the GM becomes a single Gaussian distribution.(This part is repeating here!)} 
We show estimation performance versus SNR in Fig.~\ref{fig:exp_1}. 
The results show that all NNs show bad performance when SNR is low. As the SNR improves, performance of all four NNs improves, but FFNN and ResNet outperform others. The performance of ResNet saturates at high SNR.     
%for noisier signals all the above chosen networks have similar performance. However, as the signal quality improves it is observed that the FFNN and ResNet outperforms ELM and SSFN displaying the generalising power of deeper architectures.
% \begin{Figure}
%       \includegraphics[width=1.0\textwidth,scale=1.0]{plots/A/mmse_1.eps}
% 		\captionof{figure}{Estimation performance study for visualizing role of data geometry and the resulting signal power. }
% 		%The SNR is varied by varying the scaling parameter $a$. Increase in $a$ leads to increase in distance between Gaussian components' mean vectors as shown in Figure~\ref{fig:exp_torroid_data_3d}.\textcolor{red}{I think we do not need to provide too much explanation in the caption since the text has already enough details. This will save some space.}}
% 		\label{fig:exp_1} 
% \end{Figure}

\begin{figure}[t]
	\centering
	\includegraphics[trim={0cm 0cm 0cm 0cm},clip,scale = 0.5]{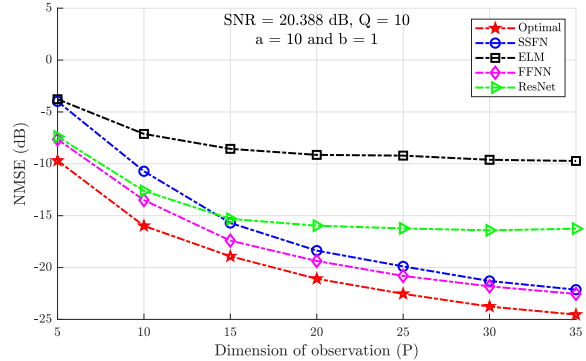}
	\caption{Estimation performance study for visualizing the role of sampling.}
	\label{fig:exp_3}
\end{figure}

The third experiment studies sampling requirements. %We choose $Q=10$, $a=10$ and $b=1$.
This is represented by the observation vector dimension $P$ for a fixed target signal dimension $Q$. Note that $P < Q$ is a under-determined sampling system, and $P \geq Q$ is an over-determined system. The NMSE versus $P$ for $Q=10$ is shown in Fig.~\ref{fig:exp_3}. The results show performance improvement as $P$ increases and then a saturation trend. 
%We again observe the saturation trend of ResNet.  
%these type of architectures may not naturally improve their performance when applied on a very higher dimensional data.
%a high sampling (a high observation vector size) helps to achieve better estimation performance. However, we also notice that while FFNN and SSFN gracefully degrade and show improving performance, ELM and ResNet have saturation trend. \textcolor{blue}{Why high dimension argument arise? 
% \begin{Figure}[t!]
%       \includegraphics[width=1.0\textwidth]{plots/A/mmse_3.eps}
% 		\captionof{figure}{Estimation performance study for visualizing the role of sampling.} %We increase the observation vector dimension $P$ from 5 to 35. Here the dimension of target signal is fixed as $Q=10$.} %The other parameters are $M=40$, $b=1$ and $a=10$.}
%         \label{fig:exp_3}
% \end{Figure}

% \begin{Figure}[b]
%       \includegraphics[width=1.0\textwidth]{plots/A/mmse_5.eps}
%         \captionof{figure}{Estimation performance study for mismatched training and testing condition. Training dataset was generated using $b=1$ that translates to SNR = 10.396 dB, and the test dataset was generated at different SNRs by varying $b$.}
%         \label{fig:exp_5}
% \end{Figure}
\begin{figure}[t]
	\centering
	\includegraphics[trim={0cm 0cm 0cm 0cm},clip,scale = 0.5]{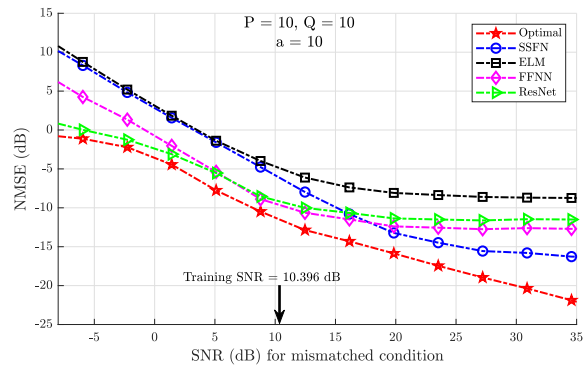}
	\caption{Estimation performance study for mismatched training and testing conditions. The training SNR is 10.396 dB.}% Training dataset was generated using $b=1$ that translates to SNR = 10.396 dB, and the test dataset was generated at different SNRs by varying $b$}
	\label{fig:exp_5}
\end{figure}
All the previous experiments used matched training and test statistics. We now consider \emph{mismatched training and test statistics} for our last experiment. A mismatched scenario is relevant in reality. In this case, different parameter settings are used to generate $\mathcal{D}_{train}$ and $\mathcal{D}_{test}$. 
%and hence the scenario is termed as \emph{mismatched training and test statistics}.
The training SNR is chosen as 10.396 dB. We vary $b$ to change test SNR keeping all other parameters same as the training condition, and then we create  $\mathcal{D}_{test}$ for a choice of $b$. The performances for different test SNRs are shown in Fig.~\ref{fig:exp_5}. 
%This experiment represent a practical scenario where we want to predict the unseen data after training the system. 
The results show that the NNs behave in a similar manner according to Fig. \ref{fig:exp_1} in the region closer to matched training and test statistics. An interesting observation is that all the four NNs show a saturation trend when test SNR increases. This result shows that training based estimators (NNs) do not generalize under different testing conditions, even when test SNR improves.  

A natural question is why ResNet-34 does not show the best performance among the competing NNs, given ResNet is known to provide a high quality performance for image classification. Our understanding is that the GMM based data that we generated, does not have similar spatial correlations like images. The convolutional filters, max-pooling and skip connections of ResNet are more suitable for image signals. Therefore it is important to understand the type and statisitical conditions of data for appropriate use and design of NNs. 

Code is available here: https://github.com/s-a-n-d-y/ExAI.

%An important observation is that a deeper network like ResNet-34 doesn't outperform the set of other methods in our experimental scenarios. To explain the behaviour of state-of-art ResNet, our guess is that the data we generated do not have spatial correlations as typically observed in images. Due to absence of such spatial correlations, it may happen that ResNet-34 is unable to capture non-linear correlation of the data which emphasizes the need of domain knowledge.   

%===================%
\section{conclusions}
%===================%

We show that it is possible to compare between NNs using generated data in a controllable experimental setup, and have an understanding of achievable performance by comparing with the MMSE bound. A large gap between performance and MMSE bound motivates effort to design efficient NNs. We conclude with an understanding that an efficient NN for a particular data type or statistical condition may not show good performance for a different data type  or statistical condition. 

\bibliographystyle{IEEEbib}
\bibliography{refs}
%==========================%

\end{document}